\pdfoutput=1

\documentclass[11pt]{article}

\usepackage{acl}
\usepackage{times}
\usepackage{latexsym}

\usepackage[T1]{fontenc}

\usepackage[utf8]{inputenc}

\usepackage{microtype}

\usepackage{inconsolata}

\usepackage{lipsum}  
\usepackage{appendix}
\usepackage{amsmath}
\usepackage{graphicx}
\usepackage{booktabs}
\usepackage{multirow}
\usepackage{color, colortbl}

\newcommand\blfootnote[1]{%
  \begingroup
  \renewcommand\thefootnote{}\footnote{#1}%
  \addtocounter{footnote}{-1}%
  \endgroup
}

\usepackage{inconsolata}
\usepackage{arydshln}

\usepackage{array} 
\newcolumntype{H}{>{\setbox0=\hbox\bgroup}c<{\egroup}@{}}

\usepackage{cancel}
\usepackage{ulem,xpatch}

\usepackage{hyperref}
\def\r{\color{red}}
\definecolor{blush}{rgb}{0.87, 0.36, 0.51}
\def\b{\color{blush}}
\definecolor{dgreen}{rgb}{0.0, 0.5, 0.0}
\def\g{\color{dgreen}}
\def\n{\color{black}}
\def\o{\color{orange}}

\usepackage{soul} 
\usepackage{inconsolata}
\usepackage{arydshln}

\usepackage{tcolorbox}
\tcbuselibrary{listings,skins,breakable}

\usepackage[colorinlistoftodos,prependcaption,textsize=tiny]{todonotes}
\definecolor{olivegreen}{RGB}{107,142,35}
\definecolor{lightolivegreen}{RGB}{157,192,105}

\newlength{\oldtabcolsep}
\usepackage{amssymb}
\def \ourmodel{Toucan}
\def \dname{AfroLingu-MT}

\title{\includegraphics[scale=0.05]{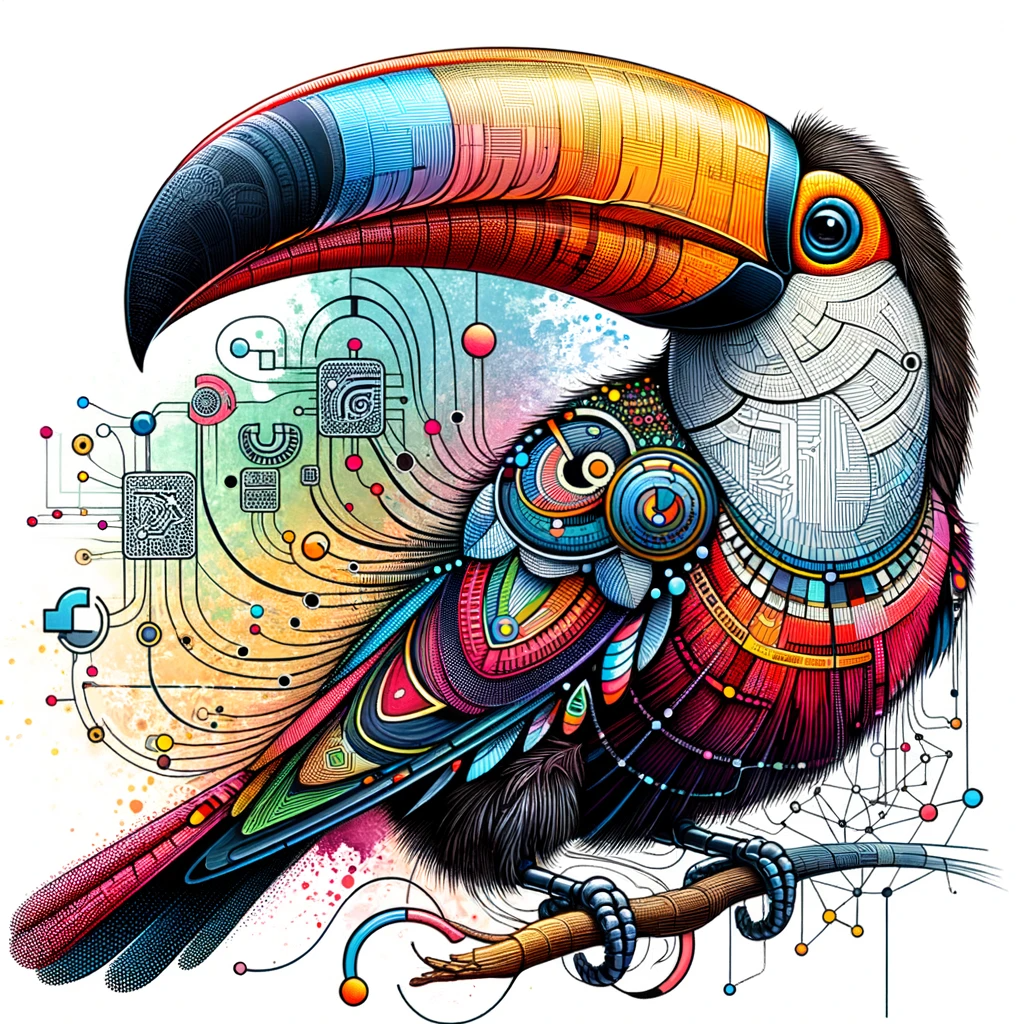}\\~\ourmodel: Many-to-Many Translation for 150 African Language Pairs}

\author{\normalsize AbdelRahim Elmadany$^{\xi,\star}$ ~ Ife Adebara$^{\xi,\star}$ ~ Muhammad Abdul-Mageed$^{\xi,\lambda}$ \\
\normalsize $^{\xi}$The University of British Columbia ~~~~~ $^\lambda$ Invertible AI\\ %
  \texttt{\normalsize \{a.elmadany,ife.adebara,muhammad.mageed\}@ubc.ca}}

\begin{document}
\maketitle

\begin{abstract}
We address a notable gap in Natural Language Processing (NLP) by introducing a collection of resources designed to improve Machine Translation (MT) for low-resource languages, with a specific focus on African languages. First, we introduce two language models (LMs), \texttt{Cheetah-1.2B} and \texttt{Cheetah-3.7B}, with $1.2$ billion and $3.7$ billion parameters respectively. Next, we finetune the aforementioned models to create \ourmodel, an Afrocentric machine translation model designed to support $156$ African language pairs. To evaluate \ourmodel, we carefully develop an extensive machine translation benchmark, dubbed \dname, tailored for evaluating machine translation. ~\ourmodel~significantly outperforms other models, showcasing its remarkable performance on MT for African languages. Finally, we train a new model, spBLEU\textsuperscript{1K}, to enhance translation evaluation metrics, covering 1K languages, including $614$ African languages. This work aims to advance the field of NLP, fostering cross-cultural understanding and knowledge exchange, particularly in regions with limited language resources such as Africa. The GitHub repository for the \textit{Toucan} project is available at \url{https://github.com/UBC-NLP/Toucan}.
\blfootnote{ $^{\star}$ Authors contributed equally. }\\
\end{abstract}
\section{Introduction}\label{sec:introduction}

Machine Translation (MT) is an important technology that bridges linguistic divides and enables communication across the globe. Although transfer learning methods~\cite{zoph-etal-2016-transfer} employing multilingual language models (mLM)~\cite{xue2021mt5, liu2020multilingual} have benefited the field specially for languages with limited resources \cite{liu-etal-2023-knn}, a significant gap remains for many African languages. In particular, although a handful of mLMs finetuned for MT for African languages~\cite{adelani-etal-2022-thousand, oladipo-etal-2023-better, jude-ogundepo-etal-2022-afriteva}, these pioneering works only serve $31$ out of the $2,000$+ languages of the African continent. This lack of coverage means the issues of language barriers, the risk of language extinction, and the under-representation of diverse communities in global conversations~\cite{koehn-knowles-2017-six}. Language barriers in particular pose significant challenges, hindering the smooth exchange of ideas, information, and cultural nuances across diverse linguistic landscapes. In contexts characterized by limited resources, where access to proficient human translators is constrained, MT has the potential to be a transformative remedy that offer unparalleled advantages in dismantling linguistic obstacles and promote heightened cross-cultural comprehension.

\begin{figure}[t]
  \centering
  \includegraphics[width=\columnwidth]{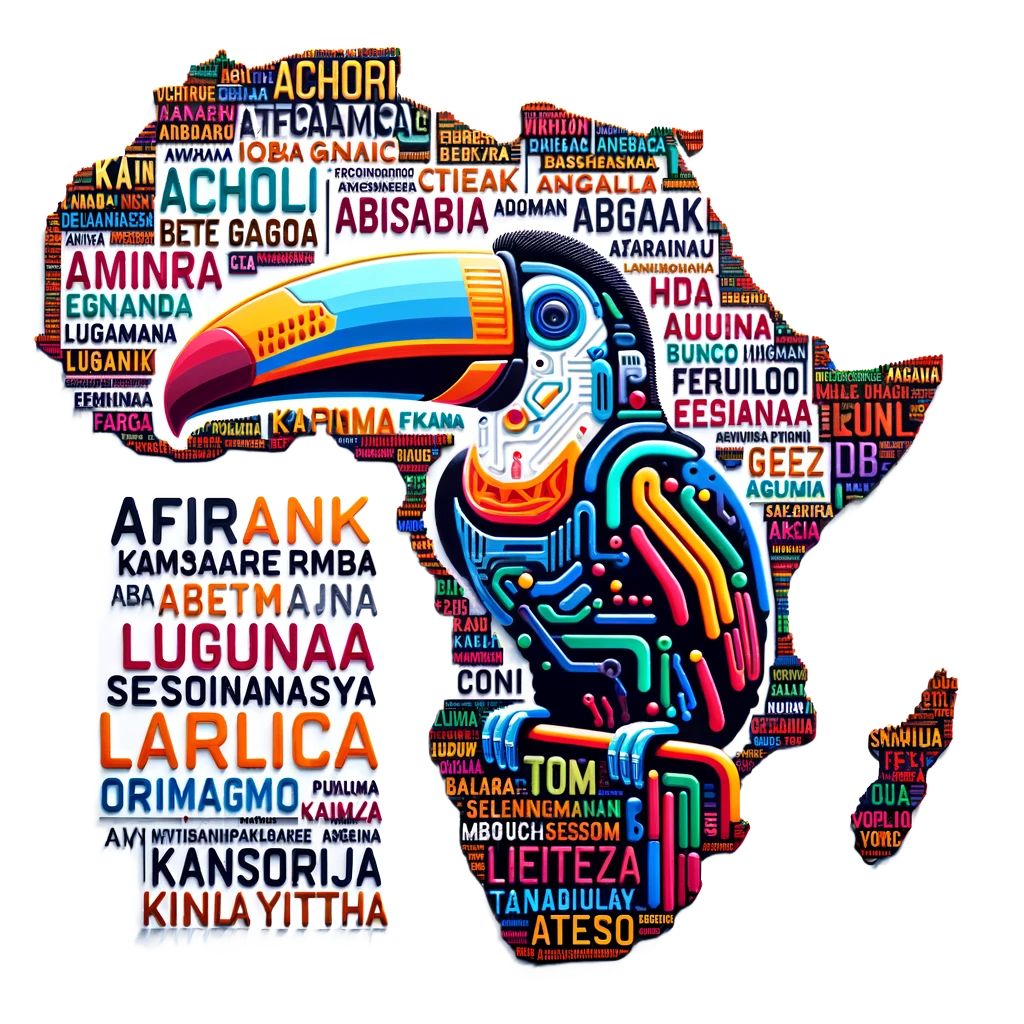}
\caption{\ourmodel~is a powerful MT model, proficiently trained on $156$ language pairs. It covers a wide spectrum of $43$ African languages as well as Arabic, English, and French}
\label{fig:countries} 
\end{figure}

In this paper, we address this gap by presenting a family of pretrained models that we also finetune for machine translation, an extensive evaluation benchmark, and an evaluation metric with wider coverage of African languages. By introducing these, we aim to contribute to the advancement of low-resources language technology especially African languages, unlocking new possibilities for cross-cultural understanding and knowledge exchange. Our main contributions can be summarized as follow:

\begin{enumerate}
    \item \noindent \textbf{\dname.} We introduce \dname, a benchmark for African languages comprising $156$ language pairs. To the best of our knowledge, \dname is the largest African MT benchmark to date. We design it to rigorously evaluate and advance the state of MT for a diverse host of African languages, addressing addressing a critical need in this area. 

    \item \noindent \textbf{Pretrained LLM.} We present a new sequence-to-sequence large language model (LLM) that covers $517$ African languages and $10$ foreign languages including - Arabic, English, French, German, Greek, Italian, Portuguese, Russian, Spanish, and Turkish. The model is available in two sizes, $1.2$B and $3.7$B parameters. We refer to these models as ``\texttt{Cheetah-$1.2$B}'' and ``\texttt{Cheetah-$3.7$B}''.

    \item \noindent \textbf{\ourmodel~models.} A versatile many-to-many family of MT models capable of translating between $46$ different languages ($43$ African languages and the three non-Indigenous major languages in Africa: Arabic, English, and French). Our models cover $156$ translation language pairs.

    \item \noindent \textbf{Comprehensive evaluation.} We offer a comprehensive comparison between generative and sequence-to-sequence LMs by evaluating a wide range of models on our \dname~benchmark under both few-shot and full finetuning scenarios. For our evaluation we use both existing models and also introduce new models as we outline next.

    \noindent \textbf{spBLEU\textsuperscript{1K}.} We introduce spBLEU\textsuperscript{1K}, a sentencepiece model that covers $1,003$ languages, including $614$ African languages, designed to improve translation evaluation quality. 
    Our model aims to address the limitations found in traditional BLEU score evaluations~\cite{peters-etal-2018-deep} and the FLORES spBLEU model~\cite{goyal2022flores} by expanding coverage to a vast array of languages that have historically been underrepresented in translation models and benchmarks.

\end{enumerate}


The rest of the paper is organized as follows: In Section \ref{sec:lit}, we discuss related work for MT benchmarks, models, and tools.  In Section~\ref{sec:mt_data} and Section~\ref{sec:toucan}, we describe \dname~evaluation benchmark and \ourmodel~models respectively. We provide details of the empirical evaluation in Section~\ref{sec:eval}, including experimental setup, baselines, and evaluation metrics. We present the results and discussion in Section \ref{sec:discussion}. Finally,  we conclude in Section \ref{sec:conc}, and outline a number of limitations, ethics and use cases for our work in Section \ref{sec:limit} and Section \ref{sec:ethics} respectively.

\section{Literature Review }\label{sec:lit}
\begin{table*}[!]
\centering
\resizebox{0.65\textwidth}{!}{%
\begin{tabular}{lll}
\toprule
\multicolumn{1}{c}{\textbf{Type}} & \textbf{Model} & \textbf{Lang/Total} \\
 \midrule
& ALMA~\cite{xu2023paradigm} & English only \\
 
& ALMA-MT~\cite{xu2023paradigm} & 0/6 \\
 
& Bloomz~\cite{muennighoff2022crosslingual}& 14/101 \\
 
& Bloomz-MT~\cite{muennighoff2022crosslingual} & Unknown/46 \\
 
& Llama-2~\cite{touvron2023llama} & English only \\ \midrule
 
\multirow{-6}{*}{CLM} & Afri-mT5~\cite{adelani-etal-2022-thousand} & 17/17 \\
 & AfriTeVa~\cite{oladipo-etal-2023-better} & 10/10 \\
 & Aya~\cite{ustun2024aya} & 15/101 \\
 & Mistral~\cite{jiang2023mistral} & English only \\
 & mT0~\cite{muennighoff2022crosslingual} & 14/101 \\
 & mT5~\cite{xue2020mt5} & 12/101 \\
\multirow{-5}{*}{Seq2Seq} & Cheetah~\cite{adebara2024cheetah} & 517/527 \\
\bottomrule
\end{tabular}%
}
\caption{Models with African languages represented. \textbf{Lang/Total}: the number of African languages covered by the model/the number of total language covered. }
\label{tab:model_comparison}
\end{table*}



MT has seen remarkable advancements, particularly in supporting underrepresented languages such as those spoken in Africa~\cite{nllbteam2022language, jude-ogundepo-etal-2022-afriteva, adelani-etal-2022-thousand}. The development of MT tools for African languages is vital for promoting linguistic diversity, fostering cross-cultural communication, and driving socio-economic progress. In this review, we briefly cover LLMs supporting African languages and related topics such as datasets and evaluation tools. Also, we provide additional details in Section~\ref{appsec:lit} in the Appendix. 
\subsection{Datasets and Benchmarks}\label{subsec:benchmarks}

A major hindrance to developing MT models for African languages is the scarcity of data~\cite{adda2016breaking, adebara-abdul-mageed-2022-towards}. While web data collection typically provides ample fine quality datasets for high-resource languages, the corpora for many African languages obtained through similar methods are often constrained in both size and quality~\cite{caswell2021quality, alabi2020massive}. To address these issues, a number of benchmarks for both training and evaluation have been developed. Notable among them are FLORES-$101$~\cite{flores-101}, FLORES-$200$~\cite{goyal2022flores}, Menyo-$20$ \cite{adelani-etal-2021-effect}, Lafand-MT \cite{adelani-etal-2022-thousand},  and Salt \cite{akera2022machine}. 

\subsection{Models and Tools}\label{subsec:lit_mt_models}
Models such as ``No Language Left Behind" (NLLB)~\cite{nllbteam2022language} cater to translation needs in $200$ languages, including $23$ African ones. Similarly, M2M-100~\cite{fan2020englishcentric} and AfriTeVa \cite{jude-ogundepo-etal-2022-afriteva} support $17$ and $10$ African languages, respectively, utilizing text-to-text architectures. AfriTeVa-V2~\cite{oladipo-etal-2023-better} has enhanced support for $16$ African languages with improved quality pretraining data. AfroMT5~\cite{adelani-etal-2022-thousand} and AfriMBART~\cite{adelani-etal-2022-thousand} each cover $17$ African languages. Additionally, Cheetah~\cite{adebara2024cheetah} extends its support to $517$ African languages and $10$ widely spoken global languages. Decoder-only models, exemplified by the Generative Pretrained Transformer (GPT)~\cite{openai2023gpt4, gpt_2020} and Llama~\cite{touvron2023llama_1, touvron2023llama}, demonstrate outstanding performance across various tasks such as text comprehension, language translation, and content generation but usually fall short in terms of coverage of African languages as we will show in this work. Aya \cite{ustun2024aya} is a massively multilingual generative language model that supports instructions following in $101$ languages including $24$ African languages. Aya has been shown to outperform models like mT0 \cite{muennighoff-etal-2023-crosslingual} and BLOOMZ \cite{muennighoff-etal-2023-crosslingual} on a wide variety of automatic and human evaluations despite covering double the number of languages.
\subsection{Evaluation Metrics}
Prominent evaluation metrics such as BLEU~\cite{papineni-etal-2002-bleu} and ChrF~\cite{popovic-2015-chrf} focus on assessing n-gram correspondence between model translations and human references, favoring precision. Meanwhile, METEOR ~\cite{banerjee-lavie-2005-meteor} emphasizes both precision and recall by considering synonyms, stemming, and word order. TER~\cite{agarwal-lavie-2008-meteor} measures edit distance between machine-generated and reference texts, while COMET~\cite{rei-etal-2020-comet, stewart-etal-2020-comet} leverages contextual embeddings for semantic similarity evaluation. These metrics aid in benchmarking machine translation (MT) systems, guiding improvements for higher quality translations.~\newcite{goyal2022flores} introduced a new metric, SentencePiece BLEU (spBLEU), which extends coverage to $101$ languages, including $23$ African languages. Similarly, AfriCOMET~\newcite{wang2023afrimte} is tailored for $17$ African languages, addressing their unique challenges in MT evaluation. These developments signify ongoing efforts to create more robust and inclusive evaluation tools for MT across diverse languages worldwide.

\section{\dname~Benchmark}\label{sec:mt_data}
In this section, we describe our data collection and construction procedures, along with the features of our comprehensive benchmark for evaluating African MT systems, \dname. To create \dname~, we execute several steps, encompassing data curation, quality evaluation, pair selection, determination of translation directions, and specification of the output format. We now explain each of these steps.

\subsection{Data Collection}
Our collection comprises data from a total of $43$ datasets, encompassing $84$ unique language pairs derived from $46$ different languages. We also develop a new manually translated dataset useful for evaluation in the government domain. In all, the data cover $43$ African languages from five language families domiciled in $29$ African countries. We also include Arabic, English, and French, since these are widely spoken in Africa. Table~\ref{tab:resources} in the Appendix provides detailed information about our collected data, including the number of pairs and data-points (i.e, examples) for each dataset. Table~\ref{tab:data_summary} (Appendix~\ref{appsec:benchmark}) on the other hand has details about each of the $46$ languages in our dataset. These tables serve as a valuable resource for understanding the breadth and depth of our datasets, ensuring transparency and facilitating further research in the field of MT for African languages. We also translate into Yoruba a portion of the Arab-acquis data \cite{habash-etal-2017-parallel} and include it in the benchmark. We refer to this data henceforth as Legal-genre. 
\begin{table}[ht!]
\scriptsize
\centering
\begin{tabular}{lrr}
\toprule
\textbf{Datasets}  & \textbf{\#Pairs} & \textbf{\#Examples} \\
\midrule
\href{https://github.com/UBC-NLP/turjuman}{AraOPUS-20} \cite{nagoudi-etal-2022-turjuman} & 3 & $2.01$M \\
\href{https://catalog.ldc.upenn.edu/LDC2016L01}{Bamanakan Lexicon}~\cite{bamanakan_2016} & $2$ &  $5,978$ \\
\href{https://github.com/AAUThematic4LT/Parallel-Corpora-for-Ethiopian-Languages}{Corpora Ethiopia}~\cite{teferra-abate-etal-2018-parallel} & $7$ &  $77,739$  \\
\href{https://zenodo.org/records/4432117\#.YjKpQInMLMa}{ENGLISH-AKUAPEM TWI}~\cite{azunre-2021-4432117} & $1$ & $25,421$ \\
\href{https://zenodo.org/records/4764039\#.YjUMVonMLMa}{English-Luganda}~\cite{mukiibi_2021_4764039} & $1$ & $15,021$ \\
\href{https://github.com/bonaventuredossou/ffr-v1/tree/master/FFR-Dataset}{FFR-Dataset} & $1$ &$136,098$ \\
\href{https://github.com/facebookresearch/flores/tree/main/flores200}{Flores-200}~\cite{nllb2022}, & $237$ & $842$ \\
\href{https://zenodo.org/records/4266935}{French-Ewe-fongbe}~\cite{degila_2020}& $2$ & $70,000$ \\
\href{https://gamayun.translatorswb.org/}{Gamayun}\cite{oktem2021congolese} & $7$  & $75,000$ \\
\href{https://opus.nlpl.eu/GlobalVoices.php}{Global Voices}~\cite{tiedemann-2012-parallel}  & $9$&  $399,178$\\
\href{https://opus.nlpl.eu/GNOME.php}{Gnome}~\cite{tiedemann-2012-parallel}  &$14$ & $850,068$ \\
\href{https://data.statmt.org/gourmet/corpora/}{Gourmet-MT}& $5$ & $263,882$     \\
\href{https://github.com/dsfsi/gov-za-multilingual/tree/master/data/sentence_align_output}{Gov-ZA}\cite{lastrucci-etal-2023-preparing, marivate_vukosi_2023_7635168} & $54$  & $638,737$ \\
\href{https://github.com/asmelashteka/HornMT/tree/main}{Horn-MT} & $15$ & $21,848$ \\
\href{https://github.com/IgnatiusEzeani/IGBONLP/tree/master}{Igbo-NLP} & $1$ & $10,008$  \\
\href{https://github.com/masakhane-io/lafand-mt/tree/main/data/tsv_files/}{Lafand-MT}~\cite{adelani-etal-2022-thousand} & $20$& $5.2M$  \\
Legal-genre  (ours) & $1$ & $3,580$ \\
\href{https://github.com/masakhane-io/masakhane-wazobia-dataset/tree/master/Data}{Masakhane Wazobia} & $2$ &$66,324$ \\
\href{https://github.com/uds-lsv/menyo-20k_MT/}{Menyo-20k}~\cite{adelani-etal-2021-effect} & $1$ &      $20,100$        \\
\href{https://opus.nlpl.eu/MultiParaCrawl.php}{Multi-paracrawl}~\cite{tiedemann-2012-parallel}& $3$& $46,478$  \\
\href{https://www.google.com/url?q=https://repo.sadilar.org\&sa=D\&source=editors\&ust=1705029413585526\&usg=AOvVaw05XRf3LCg6tNtY8hyzDJa0}{NCHLT}\cite{tiedemann-2012-parallel} & $7$ & $746,646$  \\
\href{https://www.google.com/url?q=https://opus.nlpl.eu/OpenSubtitles-v2018.php\&sa=D\&source=editors\&ust=1705029560120350\&usg=AOvVaw2HbdI8u-r-yN5kmCAI4LjQ}{Open Subtitles}~\cite{tiedemann-2012-parallel} & $1$ & $44,703$     \\
\href{https://opus.nlpl.eu/infopankki-v1.php}{Opusinfopanki}~\cite{tiedemann-2012-parallel}& $1$ & $47,220$  \\
\href{https://opus.nlpl.eu/memat.php}{Opusmemat}~\cite{tiedemann-2012-parallel}& $1$ & $139,260$   \\
\href{https://www.google.com/url?q=https://s3.amazonaws.com/web-language-models/paracrawl/bonus/}{Paracrawl} & $2$ & $147,396$    \\
\href{https://github.com/keleog/PidginUNMT/tree/master/corpus/parallel_test}{PidginUNMT\_corpus} &  $1$ & $2,101$   \\
\href{https://github.com/SunbirdAI/salt}{Salt}~\cite{akera2022machine}& $15$ & $25,000$        \\
\href{https://opus.nlpl.eu/TED2020.php}{TED}~\cite{reimers-2020-multilingual-sentence-bert}& $6$& $15,401$  \\
\href{https://opus.nlpl.eu/tico-19.php}{Tico-19}~\cite{tiedemann-2012-parallel}& $18$& $7,740$   \\
\href{https://zenodo.org/records/5035171}{Umsuka eng - zul Corpus}~\cite{rooweither_mabuya_2021_5035171} & $1$  & $10,701$   \\
\href{https://github.com/dsfsi/vukuzenzele-nlp/tree/master}{Vuk'uzenzele}~\cite{lastrucci-etal-2023-preparing, marivate_vukosi_2023_7635168} & $55$ & $66,318$ \\
\href{https://opus.nlpl.eu/wikimedia.php}{Wikimedia}~\cite{tiedemann-2012-parallel}& $21$ & $132,213$   \\
\href{https://opus.nlpl.eu/XhosaNavy.php}{Xhosanavy}~\cite{tiedemann-2012-parallel}& $1$ & $49,981$   \\
\href{https://dl.fbaipublicfiles.com/XNLI/XNLI-15way.zip}{XNLI} & $1$ & $1,000$   \\
\href{https://github.com/Niger-Volta-LTI/yoruba-text}{Yoruba Proverbs} & $1$ & $5,144$   \\
\bottomrule
\end{tabular}

\caption{Detailed description of datasets in~\dname~benchmark.}
\label{tab:resources}

\end{table} 

\begin{figure*}[!htbp] 

\begin{tcolorbox}[enhanced, breakable,
    width=\textwidth, 
    colback=white, 
    colframe=black, 
    left=1mm, right=1mm, top=1mm, bottom=1mm, 
    boxsep=1mm, 
    boxrule=0.1mm, 
    listing only, listing options={
        basicstyle=\ttfamily, 
        language=json, 
        breaklines=true 
    }]
\{"\b langcode\n":"\b nyn-ach\n", "\r instruction\n":"\r Translate the following text to Acholi language. Return only the translated sentence only. Do not repeat the instruction.\n", "\o input\n":"\o Bakakora omukoro gw'okuhendera emishomo yaabo, orwakashatu oruhwaire.\n", "\g output\n":"\g Gubed ki yub me kwero tyeko kwan i ceng adek ma okato\n"\}

\{"\b langcode\n":"\b ach-lug\n","\r instruction\n":"\r Translate the following text to Luganda language. Return only the translated sentence only. Do not repeat the instruction.\n", "\o input\n":"\o Cal pa dako man onya i nyonyo pa muno calo mac oro.\n", "\g output\n":"\g Ebifaananyi bye bibadde biyitangana ku mikuttu emigattabantu.\n"\}
\\
\{"\b langcode\n":"\b eng-teo\n","\r instruction\n":"\r Translate the following text to Ateso language. Return only the translated sentence only. Do not repeat the instruction.\n","\o input\n":"\o The Joint Anti-Terrorist Task Force car was never recovered.\n","\g output\n":"\g Mam aponi kodumunai emotoka loka Egurupu loka itunga luitijiete itunga lukodwaratau.\n"\}

\end{tcolorbox}
\caption{Examples from \dname~benchmark train set.}
\label{fig:examples}
\end{figure*}

\subsection{Data Quality}
To ensure high quality, we follow a rigorous manual review process for each dataset. This involves a thorough examination of the original paper associated with each dataset to gain a clear understanding of its data collection methodology. Following this review, we classify the datasets into four quality tiers: \texttt{Synthetic}, \texttt{Human evaluated}, \texttt{Gold}, and \texttt{Unknown}. (1) Synthetic datasets consist of translations generated solely by machine translation models, without any human quality evaluation. We exclude these datasets from further consideration and they are not part of our final collection. (2) Human evaluated quality translations are also generated by MT models, but typically undergo correction by human reviewers to improve their quality. (3) Gold quality translations are either directly translated or evaluated by human experts. Datasets in this category are also sourced from domains with a lower likelihood of containing noisy data. (4) Unknown quality datasets either lack associated publications or detailed information about their data quality and collection processes. Additionally, for certain languages, we go beyond paper analysis and conduct specific evaluations of translation quality. In particular, we manually assess translation pairs between English and languages such as Yoruba, Hausa, and Nigerian Pidgin.

\subsection{Pair Selection and Translation Directions}
We exclude ``Unknown'' and ``Syntheic'' datasets, retaining only ``Human Evaluated'' and ``Gols''  quality data and standardizing the language codes to \texttt{ISO-693}. Our objective is to facilitate development of robust MT models capable of translating between a wide range of African languages as well as  Arabic, English, and French.\footnote{Arabic, English and French are widely spoken in Africa.} To achieve this, we adopt a multifaceted approach essentially enabling many-to-many translation. For instance, the user may need to specify only the target language thus allowing for versatile translation possibilities including translation from any language into English. This results in the selection of $156$ distinct language pairs.

\subsection{Data Selection}
As depicted in Table~\ref{tab:resources} (Appendix), there is significant variation in data distribution among the language pairs, with some having a substantial number of data points and others having much fewer. To create a balanced training dataset, especially since we are targeting many-to-many translation, we aim to obtain data from each translation direction for each language pair. First, we maintain the original dataset splits where one exists from source; otherwise, we divide the language pairs into training, development, and testing datasets in an $80$/$10$/$10$ ratio. Next we sample $5$K/$50$/$200$ data points for training, development, and testing, respectively, for each language pair direction, such as English-to-Afrikaans (eng-afr) and Afrikaans-to-English (afr-eng). This approach enables us to facilitate translation for $46$ unique languages. In addition, when dealing with abundant data, we ensure there is no overlap between source and target data points for each pair in either of the directions. However, for pairs with limited data, we swap the data points between the two directions to augment the dataset. \dname~contains a total of $620,573$ parallel data points, with $586,261$ allocated for training, $7,437$ for development, and $26,875$ for the test data split. As mentioned, it comprises translations for $46$ languages derived from $156$ language pairs. We show the data distribution for each language pair in Table \ref{tab:data_statistics} (Appendix~\ref{appsec:benchmark}).

\subsection{Data Format}
Following the Alpaca style~\cite{alpaca}, we organize our dataset as follows:
   
\noindent\textbf{Lang-code}: This specifies the ISO-639-3 codes for both the source and target languages.

\noindent \textbf{Instruction}: This field provides a concise description of the task.

\noindent \textbf{Input}: This field contains the source text intended for translation.

\noindent \textbf{Output}: This includes the text translated into the target language.

We save each data point on a separate line in JSONL format. Figure~\ref{fig:examples} shows examples of translating from source language to target language in~\dname.

\begin{table}[]
\centering
\resizebox{0.96\columnwidth}{!}{%
\begin{tabular}{lc}
\toprule
\textbf{Benckmark} & \textbf{Lang/Total} \\
\midrule
FLORES-$101$~\cite{flores-101} & $23$ / $101$ \\
FLORES-$200$~\cite{goyal2022flores} & $23$ / $200$ \\
Menyo-$20$ \cite{adelani-etal-2021-effect} & $1/2$ \\
Lafand-MT \cite{adelani-etal-2022-thousand} & $16 / 18$ \\
Salt \cite{akera2022machine}& $5 / 5$\\
\midrule
\dname~(ours) & $43/46$ \\
\bottomrule
\end{tabular}%
}
\caption{\dname~benchmark (ours) in comparison with with other benchmarks with notable African language coverage. \textbf{Lang/Total} column describes the number of African languages comparing with the covered languages in the language models.}
\label{tab:compare_wz_others}
\end{table}

\subsection{\dname~in Comparison}
Table~\ref{tab:compare_wz_others} presents a comparison between~\dname~and existing benchmarks. The table highlights the total number of supported languages and language pairs in each benchmark. As Table~\ref{tab:compare_wz_others} shows, compared to other benchmarks, \dname~ doubles the number of African languages covered and has an order of magnitude higher coverage in terms of language pairs/translation directions.

\section{\ourmodel~Models}\label{sec:toucan}
In this work, we develop a number of many-to-many Afrocentric machine translation models dubbed \ourmodel. For this purpose, we first pretrain a number of Afrocentric sequence-to-sequence models that serve as the foundational backbone for our proposed machine translation \ourmodel~models.

\subsection{Cheetah Backbone LMs}
To effectively train a MT language model for African languages, it is crucial to start with a powerful, Afrocentric pretrained language model. For this purpose, we select Cheetah~\cite{adebara2024cheetah}, a recently introduced SoTA model with extensive coverage encompassing $517$ African languages. One limitation of Cheetah, however, is that it is available only in a base architecture, featuring $580$M parameters. Given our objective to develop a large-scale language model for machine translation capabale of serving 156 directions, this base model does not fully meet our requirements.

To address this limitation, we embark on training larger and more expansive Afrocentric sequence-to-sequence models. We focus on two sizes: one model with $1.2$B parameters and another with $3.7$B parameters. We refer to the new models ``\texttt{Cheetah-$1.2$B}'' and ``\texttt{Cheetah-$3.7$B}'', respectively, to reflect their enhanced capabilities and parameter scale. These models represent a significant advancement in our efforts to improve machine translation for African languages, offering greater capacities in handling the rich linguistic nuances of African languages. 
\noindent \textbf{Cheetah Pertaining}. To train the new Cheetah models, we utilize the same pretraining dataset employed in training the original Cheetah-base model~\cite{adebara2024cheetah}. This strategic choice ensures consistency in the foundational data across models, enabling the advanced Cheetah-$1.2$B and Cheetah-$3.7$B versions to build upon the rich linguistic diversity captured in the original dataset. We refer to~\cite{adebara2024cheetah} for more information about the pretraining data of Cheetah models. We employ a learning rate of $0.01$, a batch size of $1,024$ sequences, and a maximum sequence length of $1,024$. Each model undergoes pretraining for $1$ million steps. The training process is conducted on Google Cloud TPU with $128$ cores (v$3-128$) provided by the TensorFlow Research Cloud (TFRC). We provide additional details on pretraining in Section \ref{app:secmodels} in the Appendix. 

\subsection{\ourmodel~Finetuning}

We finetune the vanilla Cheetah-base model as well as the newly proposed architectures, Cheetah-$1.2$B and Cheetah-$3.7$B, on our~\dname. As explained in Section~\ref{sec:mt_data}, this dataset is the largest and most diverse African MT dataset. 
We refer to  our new models fintuned for MT as \ourmodel-base, \ourmodel-$1.2$B and \ourmodel-$3.7$B. We provide more information about model finetuning in Section~\ref{subsec:eval_setup}.

\section{Empirical Evaluation}\label{sec:eval}
\setlength{\tabcolsep}{0.5\oldtabcolsep} 

\begin{table*}[!ht]
\centering
\resizebox{\textwidth}{!}{%
\begin{tabular}{HllrccccHrccccH}
\toprule
\multicolumn{1}{c}{\multirow{2}{*}{}} & \multicolumn{1}{l}{\multirow{2}{*}{\textbf{Type}}} & \multicolumn{1}{l}{\multirow{2}{*}{\textbf{Models}}} & \multicolumn{1}{c}{\multirow{2}{*}{\textbf{\#params}}} & \multicolumn{4}{c}{\textbf{DEV}} &  & \multicolumn{4}{c}{\textbf{TEST}} \\ \cmidrule{5-9} \cmidrule{11-15}

\multicolumn{1}{c}{} & \multicolumn{1}{c}{} & \multicolumn{1}{c}{} & \multicolumn{1}{c}{} & \multicolumn{1}{c}{\textbf{\textbf{spBLEU}}} & \multicolumn{1}{c}{\colorbox{orange!20}{\textbf{spBLEU\textsuperscript{1K}}}} & \multicolumn{1}{c}{\textbf{ChrF++}} & \multicolumn{1}{c}{\textbf{AfriCOMET}} & \multicolumn{1}{H}{\colorbox{orange!20}{\textbf{AfriCOMET\textsuperscript{36}}}} &  & \multicolumn{1}{c}{\textbf{spBLEU}} & \multicolumn{1}{c}{\colorbox{orange!20}{\textbf{spBLEU\textsuperscript{1K}}}} & \multicolumn{1}{c}{\textbf{ChrF++}} & \multicolumn{1}{c}{\textbf{AfriCOMET}} & \multicolumn{1}{H}{\colorbox{orange!20}{\textbf{AfriCOMET\textsuperscript{36}}}}  \\ \midrule

\rowcolor{gray!10!} \multicolumn{15}{c}{\textbf{Zero-Shot Setting}}
\\ \midrule
 \multirow{7}{*}{   \rotatebox[origin=c]{90}{Zero-Shot}   } &\multirow{6}{*}{CLM} &ALMA 7B MT\textsuperscript{$\bigstar$} & $7$B & $1.78$ & $2.15$ & $12.3$ & $25.10$ & $0.00$ & & $1.93$ & $2.24$ & $12.31$ & $24.69$ & $0.00$ \\

 &  &Bloomz 7B MT\textsuperscript{$\bigstar$} & $7$B &  $2.02$ & $2.63$ & $11.71$ & $23.89$ & $0.00$ & & $2.29$ & $2.85$ & $12.04$ & $23.67$ & $0.00$ \\

 &  &Llama-2 7B Chat & $7$B & $1.87$ & $2.20$ & $13.42$ & $23.25$ & $0.00$ & & $1.79$ & $2.17$ & $13.29$ & $22.81$ & $0.00$ \\
 
 & & Mistral 7B Instruct v2 & $7$B &$2.00$ & $2.49$ & $14.79$ & $22.33$ & $0.00$ & & $1.89$ & $2.40$ & $14.95$ & $22.14$ & $0.00$\\
\cmidrule{3-15}

 & &ALMA 13B MT\textsuperscript{$\bigstar$} & $13$B & $2.14$ & $2.52$ & $12.24$ & $27.16$ & $0$ & & $2.00$ & $2.33$ & $12.06$ & $26.89$ & $0$\\

 & &Llama-2 13B Chat & $13$B &  $1.07$ & $1.4$ & $9.07$ & $18.67$ & $20.54$ & & $0.94$ & $1.35$ & $9.09$ & $18.2$ & $20.57$ \\

 \cmidrule{2-15}
  & S2S &mT0 XXL MT\textsuperscript{$\bigstar$} & $13$B &$5.88$ & $7.48$ & $19.79$ & $33.82$ & $0.00$ & & $6.09$ & $7.67$ & $20.09$ & $33.71$ & $0.00$ \\
   \midrule

\rowcolor{gray!10!} \multicolumn{15}{c}{\textbf{Finetuned Setting}}
\\ \midrule
  \multirow{15}{*}{ \rotatebox[origin=c]{90}{Finetuned} }& \multirow{2}{*}{CLM} &Llama-2-7B-MT & $7$B &$2.86$ & $3.39$ & $13.72$ & $24.67$ & $0.00$ & & $2.97$ & $3.54$ & $13.89$ & $24.17$ & $0.00$\\

 &   &Mistral-7B-MT & $7$B & $2.04$ & $2.61$ & $12.57$ & $20.21$ & $0.00$ & & $1.95$ & $2.57$ & $12.66$ & $21.97$ & $0.00$ \\
  \cmidrule{2-15}
& \multirow{15}{*}{S2S} &Afri-mT5 Base & $580$M & $3.23$ & $3.43$ & $16.33$ & $29.69$ & $0.00$ & & $3.28$ & $3.47$ & $16.46$ & $30.01$ & $0.00$ \\
 &  & AfriTeVa Base & $229$M &$0.00$ & $0.01$ & $8.12$ & $12.18$ & $0.00$ & & $0.00$ & $0.01$ & $8.04$ & $12.13$ & $0.00$\\
   &  & AfriTeVa V2 Base & $428$M &$3.61$ & $3.86$ & $16.94$ & $31.74$ & $0.00$ & & $3.96$ & $4.18$ & $17.45$ & $31.91$ & $0.00$\\

&  &mT0 Base& $580$M & $10.96$ & $12.12$ & $28.55$ & $48.51$ & $0.00$ & & $11.66$ & $12.88$ & $29.48$ & $48.40$ & $0.00$ \\
 &  &mT5 Base & $580$M &$11.54$ & $12.89$ & $28.89$ & $51.47$ & $0.00$ & & $11.93$ & $13.22$ & $29.46$ & $51.78$ & $0.00$\\
    &   &\includegraphics[scale=0.01]{figures/toucan_logo.png}  \ourmodel~Base (ours)   & $580$M & \underline{$16.65$} & \underline{$17.6$} & \underline{$34.09$} & \underline{$60.42$} & $0.00$ & & \underline{$17.33$} & \underline{$18.2$} & \underline{$34.56$} & \underline{$60.21$} & $0.00$ \\

 \cmidrule{3-15}

 &  &AfriTeVa Large & $745$M & $3.35$ & $3.51$ & $17.05$ & $29.41$ & $0.00$ & & $3.31$ & $3.42$ & $17.14$ & $29.25$ & $0.00$ \\ 
 &  &AfriTeVa V2 Large & $1$B &$6.06$ & $6.17$ & $22.86$ & $32.67$ & $0.00$ & & $6.24$ & $6.31$ & $23.23$ & $32.76$ & $0.00$\\
  
  &  &mT0 Large &  $1.2$B &$12.03$ & $14.94$ & $30.93$ & $50.22$ & $0.00$ & & $12.10$ & $12.01$ & $30.2$ & $50.24$ & $0.00$\\

 &  &mT5 Large &  $1.2$B &$13.28$ & $14.21$ & $30.21$ & $50.89$ & $0.00$ & & $13.33$ & $14.26$ & $30.34$ & $50.79$ & $0.00$ \\ 
 &  &\includegraphics[scale=0.01]{figures/toucan_logo.png} \ourmodel~$1.2$B (ours)& $1.2$B & $\bf 18.30$ & $\bf 19.39$ & $\bf 35.89$ & $\bf 62.58$ & $0.00$ & & $\bf 18.78$ & $\bf 19.73$ & $\bf 36.41$ & $\bf 62.36$ & $0.00$ \\

 \cmidrule{3-15}

 & & mT0 XL & 3.7B &  $14.56$ & $16.30$ & $35.54$ & $53.81$ & $0.00$ & & $14.21$ & $16.32$ & $34.34$ & $53.76$ & $0.00$ \\  
 &  &mT5 XL & 3.7B & $15.56$ & $16.03$ & $35.16$ & $53.54$ & $0.00$ & & $15.45$ & $15.95$ & $35.16$ & $53.85$ & $0.00$\\
     &  &\includegraphics[scale=0.01]{figures/toucan_logo.png} \ourmodel~$3.7$B (ours)& 3.7B & \colorbox{green!20}{$\bf 22.11$ }& \colorbox{green!20}{$\bf 22.53$} & \colorbox{green!20}{$\bf 38.91$} & \colorbox{green!20}{$\bf 66.63$} & $0.00$ & & \colorbox{green!20}{$\bf 22.67$} & \colorbox{green!20}{$\bf 23.15$} & \colorbox{green!20}{$\bf 39.53$} & \colorbox{green!20}{$\bf 66.73$} & $0.00$ \\

 \bottomrule
\end{tabular}%
}
\caption{Performance on our \dname~benchmark across both the zero-shot and full finetuning scenarios. For all causal model, we use prompting to induce translations. For sequence-to-sequence models, we use target language-based prefixes. We offer results both for development (i.e., DEV) and test (i.e., TEST) datasets. \textsuperscript{$\bigstar$}Notably, these models are trained on multilingual MT datasets. Text highlighted in \colorbox{green!20}{Bold Green} indicates the highest scores across all settings. Text highlighted in \colorbox{orange!20}{Bold Orange} indicates a new evaluation metric (ours).}\label{tab:main_results}
\end{table*}

\setlength{\tabcolsep}{\oldtabcolsep}

\setlength{\tabcolsep}{0.5\oldtabcolsep} 

\begin{table*}[!ht]
\centering
\resizebox{\textwidth}{!}{%
\begin{tabular}{HHlrccccHrccccH}
\toprule
\multicolumn{1}{H}{\multirow{2}{*}{}} & \multicolumn{1}{H}{\multirow{2}{*}{\textbf{Type}}} & \multicolumn{1}{l}{\multirow{2}{*}{\textbf{Models}}} & \multicolumn{1}{c}{\multirow{2}{*}{\textbf{\#params}}} & \multicolumn{4}{c}{\textbf{DEV}} &  & \multicolumn{4}{c}{\textbf{TEST}} \\ \cmidrule{5-9} \cmidrule{11-15}

\multicolumn{1}{c}{} & \multicolumn{1}{c}{} & \multicolumn{1}{c}{} & \multicolumn{1}{c}{} & \multicolumn{1}{c}{\textbf{\textbf{spBLEU}}} & \multicolumn{1}{c}{\textbf{spBLEU\textsuperscript{1K}}} & \multicolumn{1}{c}{\textbf{ChrF++}} & \multicolumn{1}{c}{\textbf{AfriCOMET}} & \multicolumn{1}{H}{\colorbox{orange!20}{\textbf{AfriCOMET\textsuperscript{36}}}} &  & \multicolumn{1}{c}{\textbf{spBLEU}} & \multicolumn{1}{c}{\textbf{spBLEU\textsuperscript{1K}}} & \multicolumn{1}{c}{\textbf{ChrF++}} & \multicolumn{1}{c}{\textbf{AfriCOMET}} & \multicolumn{1}{H}{\colorbox{orange!20}{\textbf{AfriCOMET\textsuperscript{36}}}}  \\ \midrule

 \multirow{7}{*}{   \rotatebox[origin=c]{90}{Zero-Shot}   } & & NLLB-$200$-$1.3$B & $1.3$B & $14.59$ & $14.74$ & $29.88$ & $53.06$ & $0$ & & $15.39$ & $15.40$ & $30.80$ & $53.16$ & $0.00$\\
 &  &\includegraphics[scale=0.01]{figures/toucan_logo.png} \ourmodel~$1.2$B (ours)& $1.2$B &  $\bf 20.83$ & $\bf 21.93$ & $\bf 38.36$ & $\bf 62.69$ & $0$ & & $\bf 21.29$ & $\bf 22.37$ & $\bf 38.80$ & $\bf 62.15$ & $0$\\

 \bottomrule
\end{tabular}%
}
\caption{A comparison between NLLB and \ourmodel~on $59$ language pairs both models support. }\label{tab:nllb_results}
\end{table*}

\setlength{\tabcolsep}{\oldtabcolsep}

\begin{table}[]
\centering
\resizebox{0.85\columnwidth}{!}{%
\begin{tabular}{lHc}
\toprule
\multicolumn{1}{l}{\textbf{Category}} & \multicolumn{1}{H}{\textbf{spBLEU}} & \multicolumn{1}{l}{\textbf{spBLEU\textsuperscript{1K}}} \\ \midrule

\rowcolor{gray!20!} Arabic $\longleftrightarrow$ XX & 16.14 & 17.27 \\
XX $\longrightarrow$  Arabic & 15.87 & 16.77 \\
Arabic $\longrightarrow$  XX & 16.41 & 17.78 \\
\rowcolor{green!20!}Arabic $\longrightarrow$  (not supported) & NA  & NA\\
\rowcolor{orange!20!}Arabic $\longrightarrow$  XX (supported) & 16.41 & 17.78 \\ \midrule
\rowcolor{gray!20!} English $\longleftrightarrow$ XX & 24.35 & 25.16 \\
XX $\longrightarrow$ English & 22.52 & 21.48 \\
English $\longrightarrow$ XX & 26.18 & 28.83 \\
\rowcolor{green!20!}English $\longrightarrow$ XX (not supported) & 28.18 & 32.00 \\
\rowcolor{orange!20!}English $\longrightarrow$ XX (supported) & 24.12 & 25.22 \\ \midrule
\rowcolor{gray!20!}French $\longleftrightarrow$ XX & 16.33 & 17.42 \\
XX $\longrightarrow$ French & 16.31 & 17.77 \\
French$\longrightarrow$ XX & 16.34 & 17.62 \\
\rowcolor{green!20!}French $\longrightarrow$ XX (not supported) & 10.61 & 12.54 \\
\rowcolor{orange!20!}French $\longrightarrow$ XX (supported) & 22.18 & 23.07 \\ \midrule
\rowcolor{gray!20!}African $\longleftrightarrow$ African & 22.99 & 22.43 \\
\rowcolor{green!20!}French $\longrightarrow$ African (not supported) & 37.71 & 38.01 \\
\rowcolor{orange!20!}French $\longrightarrow$ African (supported) & 19.45 & 18.69\\

\midrule
\rowcolor{gray!20!}Total supported languages & 26.18 & \textbf{26.57} \\
\rowcolor{gray!20!}Total unsupported languages & 21.47 & \textbf{22.36} \\

\bottomrule
\end{tabular}%
}

\caption{The performance of our \ourmodel~$3.7$B model varies based on language categories on TEST dataset. We also compare the performance between the languages in our benchmark that spBLEU supports - \colorbox{orange!20}{ $23$ supported} and \colorbox{green!20}{not supported} languages}
\label{tab:spbleu_1k}
\end{table}
\begin{table*}[!]
\scriptsize
\resizebox{\textwidth}{!}{
\begin{tabular}{lllll|lllll}
\toprule
\multicolumn{1}{c}{\multirow{2}{*}{\textbf{Lang pair}}} & \multicolumn{2}{c}{\textbf{Aya 13B}}                                      & \multicolumn{2}{c}{\textbf{\includegraphics[scale=0.01]{figures/toucan_logo.png} \ourmodel~$3.7$B (ours)}}                                  & \multicolumn{1}{c}{\multirow{2}{*}{\textbf{Lang pair}}} & \multicolumn{2}{c}{\textbf{Aya 13B}}                                      & \multicolumn{2}{c}{\textbf{\includegraphics[scale=0.01]{figures/toucan_logo.png} \ourmodel~$3.7$B (ours)}}                                  \\ \cmidrule{2-5} \cmidrule{7-10}
\multicolumn{1}{c}{}                                    & \multicolumn{1}{c}{\textbf{spBLEU}} & \multicolumn{1}{c}{\textbf{ChrF++}} & \multicolumn{1}{c}{\textbf{spBLEU}} & \multicolumn{1}{c}{\textbf{ChrF++}} & \multicolumn{1}{c}{}                                    & \multicolumn{1}{c}{\textbf{spBLEU}} & \multicolumn{1}{c}{\textbf{ChrF++}} & \multicolumn{1}{c}{\textbf{spBLEU}} & \multicolumn{1}{c}{\textbf{ChrF++}} \\ \midrule
afr→eng                             & 31.1              & 55.2             & \textbf{43.83}      & \textbf{64.63}     & eng→yor                             & 4.8               & 19.6             & \textbf{5.47}       & \textbf{23.81}     \\
amh→eng                             & \textbf{19.2}     & \textbf{44.6}    & 16.50               & 40.10              & eng→zul                             & \textbf{11.4}     & \textbf{39.7}    & 6.54                & 35.42              \\
eng→afr                             & 27.8              & 51.8             & \textbf{29.77}      & \textbf{54.61}     & hau→eng                             & 19.3              & 42.7             & \textbf{24.66}      & \textbf{47.50}     \\
eng→amh                             & \textbf{11.9}     & \textbf{23.9}    & 3.14                & 19.74              & ibo→eng                             & 16.7              & 40.3             & \textbf{19.76}      & \textbf{42.28}     \\
eng→hau                             & 11.0              & 38.4             & \textbf{18.24}      & \textbf{43.13}     & nso→eng                             & 17.3              & 40.5             & \textbf{26.88}      & \textbf{48.88}     \\
eng→ibo                             & \textbf{10.4}     & \textbf{32.3}    & 9.93                & 31.27              & sna→eng                             & 16.6              & 39.4             & \textbf{18.72}      & \textbf{41.03}     \\
eng→nso                             & \textbf{6.1}      & \textbf{29.5}    & 2.53                & 13.58              & som→eng                             & 16.8              & 40.3             & \textbf{20.65}      & \textbf{43.98}     \\
eng→sna                             & \textbf{5.6}      & \textbf{33.2}    & 0.30                & 4.80               & sot→eng                             & 20.7              & 44.2             & \textbf{26.72}      & \textbf{48.80}     \\
eng→som                             & 7.3               & 35.0             & \textbf{8.75}       & \textbf{36.28}     & swa→eng                             & 23.0              & 47.4             & \textbf{32.47}      & \textbf{56.20}     \\
eng→sot                             & \textbf{16.3}     & \textbf{42.4}    & 14.63               & 36.93              & xho→eng                             & 20.5              & 43.7             & \textbf{24.99}      & \textbf{47.42}     \\
eng→swa                             & 19.5              & 46.7             & \textbf{24.85}      & \textbf{51.16}     & yor→eng                             & 11.1              & 34.6             & \textbf{14.21}      & \textbf{36.12}     \\
eng→xho                             & \textbf{8.5}      & \textbf{36.3}    & 6.08                & 33.80              & zul→eng                             & 20.5              & 44.4             & \textbf{26.17}      & \textbf{48.69}    \\ \bottomrule
\end{tabular}}
\caption{Comparing Aya 13.B with Toucan on their intersection of pairs. Toucan outperforms Aya on $16 $ of $28$ pairs. Comparison is done with Flores devtest splits.}\label{tab:aya_toucan_results}
\end{table*}

\subsection{Evaluation Settings}\label{subsec:eval_baselines}

We evaluate \dname~in diverse scenarios, including both finetuning and zero-shot settings. \textbf{(1)} We conduct comparative analyses between \textbf{multilingual and Africa-centric pretrained language models by finetuning} these models on our~\dname~dataset. Specifically, we utilize \texttt{mT5}~\cite{xue2020mt5} and \texttt{mT0} ~\cite{muennighoff2022crosslingual} as our representative multilingual pretrained models. In contrast, we compare these with African-specific models such as \texttt{Afri-mT5}~\cite{adelani-etal-2022-thousand}, \texttt{AfriTeVa}~\cite{oladipo-etal-2023-better}, and \texttt{Cheetah}~\cite{adebara2024cheetah}. \textbf{(2)} We also evaluate the performance of \textbf{instruction-following LLMs on \dname~in a zero-shot} setting, employing a prompt-based technique. We utilize LLMs that have been trained on general instructions, such as \texttt{LLaMA-2}~\cite{touvron2023llama}, \texttt{Mistral}~\cite{jiang2023mistral}, and \texttt{ALMA}~\cite{xu2023paradigm}. Additionally, we assess LLMs that have been specifically trained on machine translation data, including \texttt{Bloomz-MT}~\cite{muennighoff2022crosslingual} and \texttt{mT0-XXL-MT}~\cite{muennighoff2022crosslingual}. Table~\ref{tab:model_comparison} shows the comparison between these models based on the African languages represented.

\subsection{Experimental Setup}\label{subsec:eval_setup}
We finetune all models on \dname~for $5$ epochs. For base and large architectures, we finetune the models using a learning rate of \texttt{5e\textsuperscript{-5}}, a batch size of $8$, and a maximum sequence length of $512$ tokens. For XL model (3.7B parameters), we use a learning rate \texttt{2e\textsuperscript{-5}}, a batch size of $2$, and a maximum sequence length of $256$ tokens. During the training process, we implement a linear learning rate scheduler, incorporating a warm-up phase that accounts for $10$\% of the total training steps.


In all finetuning experiments, we rigorously select the best-performing checkpoint for each model based on performance in the respective development set. We then report and analyze the performance of each model on the corresponding test set.

\subsection{Evaluation Metrics}\label{subsec:eval_metrics}
In this work, we present the performance outcomes of our proposed models as well as the baseline models each evaluated independently on the \dname~benchmark. This evaluation employs three pertinent metrics specific to machine translation. These metrics are:  SentencePiece BLEU (i.e., \texttt{spBLEU})~\cite{goyal2022flores},  word-based Character n-gram F-score (i.e., \texttt{ChrF++})~\cite{popovic2015chrf}, and \texttt{AfriCOMET}~\cite{wang2023afrimte}. These metrics have been selected for their effectiveness in assessing the quality of machine translations from various perspectives, including lexical accuracy and fluency. We also introduce a wider coverage metric modeled after spBLEU, dubbed spBLEU\textsuperscript{1K}, as we explain next.

\subsubsection{spBLEU\textsuperscript{1K}}
Employing the BLEU metric for evaluating translations, particularly in the context of low-resource languages, is suboptimal due to its fundamental reliance on n-gram overlap. This reliance significantly impacts the metric's effectiveness, as it is heavily influenced by the specific tokenization method used. Notably, employing a more aggressive tokenization strategy can lead to artificially inflated BLEU scores~\newcite{goyal2022flores}. To address this,~\newcite{goyal2022flores} proposed a novel metric, SentencePiece BLEU (i.e., \texttt{spBLEU}), designed to measure and analyze the performance of translations across $101$ languages. This approach involves training a new SentencePiece-based tokenizer using monolingual data for $101$ languages, replacing the default tokenizer typically used in SacreBLEU, known as `mosestokenizer' \cite{post-2018-call}. This innovation aims to standardize the tokenization process, thus providing more accurate and comparable translation performance metrics across many languages.

Significantly, the spBLEU metric covers merely $23$ out of the $43$ languages present in our \dname~benchmark. To address this limitation, we adopt a methodology similar to that of~\newcite{goyal2022flores}. Namely, we develop a new SentencePiece tokenizer that utilizes $1000$+ monolingual data sources.

\noindent \textbf{Data}. We collect monolingual data covering $1,003$ languages, including $614$ African languages, $53$ Indigenous American languages, and the remainder spanning the most resource-rich languages worldwide. We use a diverse data source, encompassing Wikipedia, Wikibooks, the Bible, newspapers, and common web sources. Additionally, we utilize the MADLAD dataset~\cite{kudugunta2023madlad400}, which covers $419$ languages. A list with the $1,003$ we cover is available at \href{https://github.com/UBC-NLP/Toucan}{Toucan}.

\noindent \textbf{Training a SentencePiece Model (SPM)}. One significant challenge is the uneven availability of monolingual data across various languages. This disparity is especially acute for low-resource languages, which often suffer from a lack of comprehensive coverage in subword units and may not possess a sufficiently large corpus of sentences to ensure a broad and diverse representation of content. To address this issue and enhance the training of our new SPM, we adopt a temperature upsampling technique similar to the methodology described in ~\newcite{conneau2019unsupervised}.

\noindent \textbf{Integrating with SacreBleu}. We integrate this newly created SPM into SacreBLEU, resulting in the formulation of our more inclusive metric spBLEU\textsuperscript{1K}. Our metric is thus designed to provide a more comprehensive evaluation across a broader range of languages, including those that are underrepresented in existing metrics such as spBLEU.

\section{Results and Discussion}\label{sec:discussion}

\noindent \textbf{spBLEU\textsuperscript{1K} Metric}. The results indicate that our new metric, spBLEU\textsuperscript{1K}, enhances the translation scores by an average of $0.74$ and $0.62$ BLEU points on the development and test datasets, respectively. We also evaluate the performance of our best model, \ourmodel-$3.7$B, based on language categories. Table~\ref{tab:spbleu_1k} presents a comparison between spBLEU and our new evaluation metric across different categories of translation directions in our data. Results demonstrate that the two metrics are almost identical in the shared languages (the languages that both metrics supported), however, the new metric improves the translation results in languages not supported by spBLEU. 

Exploring different pretraining settings allows us to derive unique insights. Examples of insights that can be gleaned from Table~\ref{tab:main_results} include:\\

\noindent \textbf{Sequence-to-sequence models with wider coverage perform better MT.} Unsurprisingly, the findings show that sequence-to-sequence models pretrained with more languages enable better MT performance on our dataset. For example, \ourmodel, which supports $517$ African languages and ten of the most spoken languages worldwide, outperforms all other models. On the other hand, AfriTeVa (which is pretrained on only ten languages) has the lowest performance. However, AfriTeVa-v2 outperforms Afri-MT5 even though it supports fewer languages. We assume that the size of the LM may plays a role here.  \\

\noindent \textbf{Models of the same size finetuned on more language pairs tend to perform better translations.} Among the models we evaluate in zero-shot, four are already finetuned on external MT parallel data. These are ALMA-7B-MT, Bloomz-7B-MT, ALMA-13B-MT, and MT0-XXL-MT (13B). While we do not see a clear pattern for the 7B model size, our results for the 13B do carry a pattern: Results show that the MT0-XXL-MT model (13B), which is finetuned on pairs from 46 languages gives significantly better translations than ALMA-13B-MT, which is finetuned on pairs from six languages (i.e., 5.34 points in spBLEU\textsuperscript{1k} on TEST).  \\



\noindent \textbf{Larger models perform better}. Again, unsurprisingly, we observe that larger models outperform smaller ones. For example, models with $1.2$B parameters outperform base models by approximately \textit{two} points spBLEU\textsuperscript{1k}, whereas the larger models with $3.7$B outperform both the base and $1.2$B parameter models by approximately \textit{five} and \textit{three} points spBLEU\textsuperscript{1k}, respectively. Additionally, we note that our model, \ourmodel, outperforms the second best performing model (i.e., mT5) base, $1.2$B, and $3.7$B models by $4.98$, $7.47$, and $7.2$ points spBLEU\textsuperscript{1k} on TEST, respectively.

Additionally, we compare our model, \ourmodel-$1.2$B, to the Facebook's NLLB model \cite{nllbteam2022language, nllb2022}. Again, we find \ourmodel-1.2B outperforming NLLB-$200$-$1.3$B by $6.96$ points in spBLEU\textsuperscript{1K}, as shown in Table~\ref{tab:nllb_results}. \\

\noindent \textbf{Toucan outperforms Aya}
We compare the performance of Toucan with Aya \cite{ustun2024aya}. We use results for Aya as they appear in the paper, hence, we do not compute spBLEU\textsuperscript{1k} results in this analysis. Although Aya is a 13B parameter model, significantly larger than Toucan 3.7B, we report better performance in $16$ of $28$ pairs. In Table \ref{tab:aya_toucan_results}, we show the performance of Toucan and Aya on Flores200.


\section{Conclusion}\label{sec:conc}
In this paper, we introduce a suite of resources aimed at enhancing MT for low-resource African languages. We present \texttt{Cheetah-1.2B} and \texttt{Cheetah-3.7B}, with 1.2B  and 3.7B parameters, respectively. We further finetune these models into versatile many-to-many model capable of translating between $46$ different languages, including $43$ African languages and the three major non-Indigenous languages in Africa: Arabic, English, and French. We also introduce \dname~as the largest African MT benchmark to date. We provide a comprehensive comparison between various LLMs by evaluating a wide range of models on our \dname~benchmark. Finally we extend spBLEU - spBLEU\textsuperscript{1K} - a sentencepiece-based evaluation metric covering $1,003$ languages, including $614$ African languages. This aims to enhance translation evaluation quality, particularly for languages historically underrepresented in translation models and benchmarks. 
\section{Limitations}\label{sec:limit}

We can identify a number of limitations that are relevant to our work, as follows:

\begin{itemize}
\item Even though we cover the largest number of African languages in our MT models, compared to previous research, there is still a long list of African languages that lack MT support. It will take the community more work to develop new parallel datasets for these languages so that they can be supported. We believe that our new Cheetah models can be valuable in this regards, since they have a coverage of $517$ languages and can be easily finetuned on new languages once parallel datasets are available. 
  
\item Another limitation is that metrics such as AfriCOMET can only cover $17$ African languages, although we developed the spBLEU\textsuperscript{1K} metric that allowed us to evaluate on our dataset in a dependable way since it is based on wide coverage vocabulary. Again, new parallel datasets can be helpful for extending COMET to more African languages.

\end{itemize}

\section{Ethics Statement and Wider Impacts}\label{sec:ethics}

Our model, \ourmodel, is rooted in Afrocentric NLP principles, prioritizing the technological needs of African communities. We anticipate that \ourmodel~will not only benefit speakers of the supported languages but also aid researchers in African languages, including anthropologists and linguists. Below, we highlight some potential use cases for \ourmodel~and discuss its broader impacts:

Addressing the technological disparity in approximately 90\% of the world's languages, which disproportionately affects native speakers, \ourmodel~focuses specifically on Africa. As the first massively multilingual pre-trained language model (PLM) developed for African languages and their variants, it encompasses knowledge of $517$ African languages, making it the largest model for African NLP to date.

\ourmodel~facilitates enhanced access to critical information for the African community through Indigenous African languages. Particularly beneficial for individuals with limited proficiency in other languages, this capability has the potential to foster greater global connectivity.

\ourmodel~supports language preservation efforts for numerous African languages, many of which have not been utilized in NLP tasks before. We believe it can encourage the continued use of these languages across various domains and spur the development of language technologies tailored to their specific needs.

Despite their versatility, language models like \ourmodel~can be susceptible to misuse. Developed using publicly available datasets that may contain biases, we strive to conduct analyses and diagnostic case studies to assess our model's performance. However, our investigations are not exhaustive and do not guarantee the absence of bias in the data, especially considering limited access to native speakers of most covered languages.

In summary, \ourmodel~represents a significant step forward in Afrocentric NLP, addressing technological disparities, fostering language preservation, and promoting responsible deployment of language technologies in African contexts.

\normalem

\bibliography{UBC_African_anthology}
\appendix

\clearpage
\appendixpage
\addappheadtotoc

\numberwithin{figure}{section}
\numberwithin{table}{section}


     



\section{Literature Review }\label{appsec:lit}

The recent years have witnessed a substantial evolution in LLMs, marked by numerous breakthroughs across various applications such as MT and text summarization. Particularly in MT, significant advancements have occurred, with a notable emphasis on supporting underrepresented languages, including those spoken in Africa \cite{nllbteam2022language, jude-ogundepo-etal-2022-afriteva, adelani-etal-2022-thousand}. The development of MT tools tailored for African languages holds significant importance in promoting linguistic diversity, facilitating cross-cultural communication, and driving socio-economic progress within the region \cite{adebara2024cheetah}. 

In this section, we explore the current landscape of LLMs supporting African languages, covering topics such as datasets and benchmarks, instruction fine-tuning, evaluation metrics, challenges, and practical applications. It examines key contributions and advancements within this dynamic field, providing insights into its ongoing development and potential impact.


\noindent\textbf{Datasets and Benchmarks}
One of the primary hindrances to developing MT models for African languages is the scarcity of data \cite{adda2016breaking, adebara-abdul-mageed-2022-towards}. For many high resource languages, collecting data from the web often yields large and high quality datasets. However, resulting corpora for many African languages using similar methods are often limited in size and quality \cite{caswell2021quality, alabi2020massive}. Furthermore, texts from religious domains dominate the data landscape with most datasets coming from bibles \cite{mccarthy-etal-2020-johns} and other religious documents \cite{agic-vulic-2019-jw300}. To address these issues, a handful of benchmarks for both training and evaluation have been developed. Notable among them are FLORES-$101$~\cite{flores-101}, FLORES-$200$~\cite{goyal2022flores}, Menyo-$20$ \cite{adelani-etal-2021-effect}, Lafand-MT \cite{adelani-etal-2022-thousand},  and Salt \cite{akera2022machine}. 

\noindent \textbf{Models and Tools} A few African languages have benefited from the recent advancement of LM. We now describe models and tools that support African languages. (1) No Language Left Behind (NLLB) is a suite of open-sources models capable of delivering evaluated, high-quality translations directly between $200$ languages including $23$ African languages \cite{nllbteam2022language}. (2) M2M-100 supports $100$ languages including $17$ African languages \cite{fan2020englishcentric}. (3) AfriTeVa~\cite{jude-ogundepo-etal-2022-afriteva} supports ten African languages using a T5-style model. (4) AfriTeVa V2~\cite{oladipo-etal-2023-better} supports $16$ African languages using a better quality pre-training data. (5) AfroMT5~\cite{adelani-etal-2022-thousand} and AfriMBART \cite{adelani-etal-2022-thousand} each support $17$ African languages. (6) T5 (Text-To-Text Transfer Transformer) based models such as mT5~\cite{xue-etal-2021-mt5}, mT0 and \cite{muennighoff2022crosslingual}. (7) Cheetah~\cite{adebara2024cheetah}, supports $517$ African languages and $10$ widely spoken languages in the world. 


Decoder-only models have also shown remarkable improvements across multiple tasks. Now, we describe some of these modes. (1) Generative Pretrained Transformer (GPT)~\cite{openai2023gpt4, gpt_2020}, a transformer architecture model, pretrained on expansive datasets, resulting in its proficiency in generating coherent, contextually rich text. (2) Llama~\cite{touvron2023llama_1, touvron2023llama} represents a state-of-the-art language model designed for nuanced natural language understanding and generation. LLama excels in tasks ranging from comprehensive text comprehension to intricate language translation and sophisticated content creation. Its versatility positions it as a pivotal tool within the realm of artificial intelligence, particularly in contexts requiring nuanced linguistic analysis. These models exhibit prowess in various facets of natural language processing, treating each task as a text-to-text challenge. Its capabilities span translation, summarization, and question answering, showcasing versatility and effectiveness.


\noindent \textbf{Evaluation Metrics} MT evaluation metrics are tools used to assess the quality of translations generated by MT models/systems. These metrics provide objective, quantifiable measures of how accurately and fluently a machine-translated text matches a reference translation, typically produced by humans. Among the most widely used BLEU~\cite{papineni-etal-2002-bleu}, ChrF~\cite{popovic-2015-chrf} evaluate the correspondence of n-grams between the machine-generated text and the reference, favoring precision. METEOR~\cite{banerjee-lavie-2005-meteor}, on the other hand, emphasizes both precision and recall, incorporating synonyms, stemming, and word order into its evaluation. TER \cite{agarwal-lavie-2008-meteor} measures the number of edits required to change a machine-translated text into a reference text. COMET~\cite{rei-etal-2020-comet, stewart-etal-2020-comet} leverage contextual embedding to capture semantic similarities more effectively. These metrics facilitate the benchmarking of MT systems, guiding researchers and developers in improving translation technologies to achieve higher quality and more natural translations.

Recently, researchers have made strides in enhancing the capabilities of evaluation metrics like BLEU and COMET to better accommodate low-resource languages. ~\newcite{goyal2022flores} introduced a novel metric, SentencePiece BLEU (spBLEU), which extends coverage to $101$ languages, including $23$ African languages. This development marks a significant step forward in making language technology more inclusive, especially for languages that have traditionally been underrepresented in machine translation research. Similarly,~\newcite{wang2023afrimte} developed AfriCOMET, a COMET-based evaluation metric, specifically designed to support $17$ African languages. AfriCOMET represents another leap towards recognizing and addressing the unique challenges posed by African languages in machine translation. Both spBLEU and AfriCOMET exemplify the ongoing efforts within the research community to develop more robust and inclusive tools for evaluating machine translation quality across a broader spectrum of the world's languages.

\section{Cheetah Models Pretraining Details}\label{app:secmodels}
\noindent\textbf{Vocabulary.} We employ SentencePiece \cite{kudo-richardson-2018-sentencepiece} to encode text into WordPiece tokens \cite{sennrich-etal-2016-neural}, utilizing $250$K WordPieces. Additionally, our dataset encompasses the top ten globally spoken languages: Arabic, English, French, German, Greek, Italian, Portuguese, Russian, Spanish, and Turkish, sourced from Wikipedia dumps. Each language comprises 1 million sentences, solely included in the vocabulary.

\noindent \textbf{Models Architecture.} We pretrain \dname using the encoder-decoder architecture \cite{xue-etal-2021-mt5}. Both the encoder and decoder components are structured similarly to T5, featuring $12$ layers, each with $12$ attention heads, and $768$ hidden units for the base model. Consequently, the model comprises approximately $\sim580$ million parameters.

\noindent{\textbf{Objective.}} We employ an unsupervised (denoising) objective, where the model is trained on masked (corrupted) versions of the original sentence to reconstruct the original sequence \cite{xue-etal-2021-mt5}. This objective involves randomly sampling and dropping out 15\% of tokens in the input sequence. Subsequently, all consecutive spans of dropped-out tokens are replaced by a single sentinel token.

\noindent \textbf{Pretraining Procedure} During the pretraining of \dname, we employ a learning rate of $0.01$, a batch size of $1,024$ sequences, and a maximum sequence length of $1,024$. Each model undergoes pretraining for $1$ million steps. The training process is conducted on Google Cloud TPU with $128$ cores (v$3-128$) provided by the TensorFlow Research Cloud (TFRC).\footnote{\href{https://sites.research.google/trc/about/}{https://sites.research.google/trc/about/}}

\section{\dname~Benchmark}\label{appsec:benchmark}
\begin{table*}[!ht]
\centering
\resizebox{\textwidth}{!}{%
\begin{tabular}{llllll}
\toprule
\textbf{ISO}                & \textbf{Name}                    & \textbf{Country(ies)}   
& \textbf{\# of Speakers} & \textbf{Family} & \textbf{Script} \\
\midrule
aar                         & Afar                             & Ethiopia                                    & 2.36M                   & Afro-Asiatic    & Latin           \\
ach                         & Acholi                           & Uganda, South Sudan                         & 1.58M                   & Nilo-Saharan    & Latin           \\
afr                         & Afrikaans                        & South Africa                                & 17.67M                  & Indo-European   & Latin           \\
aka                         & Akan                             & Ghana                                       & 9.88M                   & Niger-Congo     & Latin           \\
amh                         & Amharic                          & Ethiopia                                    & 57.56M                  & Afro-Asiatic    & Ethiopic        \\
ara                         & Arabic                           & North Africa                                & 372.56K                 & Afro-Asiatic    & Arabic          \\
bam                         & Bambara                          & Côte d’Ivoire, Mali                         & 14.18M                  & Niger-Congo     & Latin           \\
bas                         & Bassa                            & Cameroon                                    & 300K                    & Niger-Congo     & Latin           \\
bem                         & Bemba                            & Zambia, Democratic Republic of Congo        & 4.11M                   & Niger-Congo     & Latin           \\
btg                         & Bhete                            & Côte d’Ivoire                               & 329K                    & Niger-Congo     & Latin           \\
eng & English                          & Global                                      & 1.45B                   & Indo-European   & Latin           \\
ewe                         & Ewe                              & Ghana, Togo                                 & 5.5M                    & Niger-Congo     & Latin           \\
fan                         & Fang                             & Equatorial Guinea, Cameroon, Gabon          & 1.06M                   & Niger-Congo     & Latin           \\
fon                         & Fon                              & Benin, Togo                                 & 2.28M                   & Niger-Congo     & Latin           \\
fra & French                           & Global                                      & 310M                    & Indo-European   & Latin           \\
gez                         & Ge'ez                            & Ethiopia                                    & Religious use           & Afro-Asiatic    & Ethiopic        \\
hau                         & Hausa                            & Nigeria                                     & 78.52M                  & Afro-Asiatic    & Latin           \\
ibo                         & Ibo                              & Nigeria                                     & 30.89M                  & Niger-Congo     & Latin           \\
kau                         & Kanuri                           & Nigeria, Niger                              & 9.47M                   & Nilo-Saharan    & Latin           \\
kbp                         & Kabiye                           & Togo, Benin                                 & 992K                    & Niger-Congo     & Latin           \\
kin                         & Kinyawanda                       & Rwanda                                      & 14.52M                  & Niger-Congo     & Latin           \\
kon                         & Kongo                            & Democratic Republic of Congo, Angola, Congo & 7.01M                   & Niger-Congo     & Latin           \\
lgg                         & Lugbara                          & Uganda, Democratic Republic of Congo        & 1.94M                   & Nilo-Saharan    & Latin           \\
lin                         & Lingala                          & Democratic Republic of Congo, Congo         & 40.27M                  & Niger-Congo     & Latin           \\
lug                         & Luganda                          & Uganda                                      & 11.01M                  & Niger-Congo     & Latin           \\
mlg                         & Malagasy                         & Madagascar                                  & 25M                     & Austronesian    & Latin           \\
nnb                         & Nande                            & Democratic Republic of Congo                & 903K                    & Niger-Congo     & Latin           \\
nya                         & Chichewa                         & Malawi, Mozambique, Zambia                  & 13.38M                  & Niger-Congo     & Latin           \\
nyn                         & Nyankore                         & Uganda, Rwanda                              & 3.43M                   & Niger-Congo     & Latin           \\
orm                         & Oromo                            & Ethiopia                                    & 37.45M                  & Afro-Asiatic    & Latin           \\
pcm                         & Nigerian Pidgin                  & Nigeria                                     & 116M                    & Creole          & Latin           \\
som                         & Somali                           & Somalia                                     & 22.04M                  & Afro-Asiatic    & Latin           \\
sot                         & Sesotho                          & Lesotho                                     & 13.52M                  & Niger-Congo     & Latin           \\
ssw                         & Siswati                          & Eswatini                                    & 4.71M                   & Niger-Congo     & Latin           \\
swa                         & Swahili                          & Kenya, Tanzania                             & 73M                     & Niger-Congo     & Latin           \\
swc                         & Swahili Congo                    & Democratic Republic of Congo                & 11.14M                  & Niger-Congo     & Latin           \\
teo                         & Ateso                            & Uganda, Kenya                               & 2.77M                   & Nilo-Saharan    & Latin           \\
tir                         & Tigrinya                         & Eritrea, Ethiopia                           & 8.82M                   & Afro-Asiatic    & Ethiopic        \\
tsn                         & Tswana                           & Botswana                                    & 13.75M                  & Niger-Congo     & Latin           \\
tso                         & Tsonga                           & South Africa                                & 10M                     & Niger-Congo     & Latin           \\
twi                         & Twi                              & Ghana                                       & 9.88M                   & Niger-Congo     & Latin           \\
wal                         & Wolaytta & Ethiopia                                    & 2.49M                   & Afro-Asiatic    & Latin           \\
wol                         & Wolof                            & Senegal, Mauritania                         & 12.39M                  & Niger-Congo     & Latin           \\
xho                         & Xhosa                            & South Africa                                & 19.21M                  & Niger-Congo     & Latin           \\
yor                         & Yoruba                           & Nigeria                                     & 45.86M                  & Niger-Congo     & Latin           \\
zul                         & Zulu                             & South Africa                                & 27.8M                   & Niger-Congo     & Latin           \\
\bottomrule
\end{tabular}%
}
\caption{Details of the Languages in our dataset.}\label{tab:data_summary}
\end{table*}
\begin{table*}[!ht]
\scriptsize
\centering
\resizebox{\textwidth}{!}{%
\begin{tabular}{lrrr|lrrr|lrrr}
\toprule
\textbf{Lang Pair} & \multicolumn{1}{l}{\textbf{Train}} & \multicolumn{1}{l}{\textbf{Dev}} & \multicolumn{1}{l}{\textbf{Test}} & \textbf{Lang Pair} & \multicolumn{1}{l}{\textbf{Train}} & \multicolumn{1}{l}{\textbf{Dev}} & \multicolumn{1}{l}{\textbf{Test}} & \textbf{Lang Pair} & \multicolumn{1}{l}{\textbf{Train}} & \multicolumn{1}{l}{\textbf{Dev}} & \multicolumn{1}{l}{\textbf{Test}} \\
\midrule
aar-amh            & 1166                               & 50                               & 145                               & eng-som                                    & 5,000                              & 50                               & 200                               & nyn-eng                                    & 5000                               & 50                               & 200                               \\
aar-eng            & 1146                               & 50                               & 143                               & eng-sot                                    & 4408                               & 96                               & 248                               & nyn-lgg                                    & 5000                               & 50                               & 200                               \\
aar-orm            & 1166                               & 50                               & 145                               & eng-ssw                                    & 580                                & 36                               & 72                                & nyn-lug                                    & 5000                               & 50                               & 200                               \\
aar-som            & 1166                               & 50                               & 146                               & eng-swa                                    & 5000                               & 50                               & 200                               & nyn-teo                                    & 5000                               & 50                               & 200                               \\
aar-tir            & 1165                               & 50                               & 145                               & eng-teo                                    & 5000                               & 50                               & 200                               & orm-aar                                    & 1166                               & 50                               & 145                               \\
ach-eng            & 5000                               & 50                               & 200                               & eng-tir                                    & 5000                               & 50                               & 200                               & orm-amh                                    & 5000                               & 50                               & 200                               \\
ach-lgg            & 5000                               & 50                               & 200                               & eng-tsn                                    & 5000                               & 50                               & 200                               & orm-eng                                    & 5000                               & 50                               & 200                               \\
ach-lug            & 5000                               & 50                               & 200                               & eng-tso                                    & 5000                               & 50                               & 200                               & orm-som                                    & 1170                               & 50                               & 146                               \\
ach-nyn            & 5000                               & 50                               & 200                               & eng-twi                                    & 5000                               & 50                               & 200                               & orm-tir                                    & 4980                               & 50                               & 200                               \\
ach-teo            & 5000                               & 50                               & 200                               & eng-wal                                    & 5000                               & 50                               & 200                               & orm-wal                                    & 2338                               & 50                               & 146                               \\
afr-eng            & 5000                               & 50                               & 200                               & eng-wol                                    & 5000                               & 50                               & 200                               & pcm-eng                                    & 1681                               & 50                               & 105                               \\
aka-eng            & 3145                               & 50                               & 196                               & eng-xho                                    & 5000                               & 50                               & 200                               & som-aar                                    & 1166                               & 50                               & 146                               \\
amh-aar            & 1166                               & 50                               & 145                               & eng-yor                                    & 5000                               & 50                               & 200                               & som-amh                                    & 1156                               & 50                               & 145                               \\
amh-eng            & 5000                               & 50                               & 200                               & eng-zul                                    & 5000                               & 50                               & 200                               & som-eng                                    & 5000                               & 50                               & 200                               \\
amh-gez            & 4618                               & 50                               & 200                               & ewe-eng                                    & 128                                & 16                               & 16                                & som-fra                                    & 4231                               & 50                               & 200                               \\
amh-mlg            & 693                                & 43                               & 87                                & ewe-fra                                    & 5000                               & 50                               & 200                               & som-orm                                    & 1170                               & 50                               & 146                               \\
amh-orm            & 5000                               & 50                               & 200                               & fan-btg                                    & 222                                & 28                               & 27                                & som-swa                                    & 1390                               & 50                               & 173                               \\
amh-som            & 1156                               & 50                               & 145                               & fon-fra                                    & 5000                               & 50                               & 200                               & som-tir                                    & 1153                               & 50                               & 144                               \\
amh-swa            & 112                                & 14                               & 14                                & fra-ara                                    & 5000                               & 50                               & 200                               & sot-eng                                    & 4408                               & 96                               & 248                               \\
amh-tir            & 5000                               & 50                               & 200                               & fra-ewe                                    & 5000                               & 50                               & 200                               & ssw-eng                                    & 580                                & 37                               & 72                                \\
amh-wal            & 3763                               & 50                               & 200                               & fra-fon                                    & 5000                               & 50                               & 200                               & swa-amh                                    & 112                                & 14                               & 14                                \\
ara-eng            & 5000                               & 50                               & 200                               & fra-lin                                    & 4000                               & 50                               & 200                               & swa-eng                                    & 5000                               & 50                               & 200                               \\
ara-fra            & 5000                               & 50                               & 200                               & fra-nnb                                    & 5000                               & 50                               & 200                               & swa-fra                                    & 5000                               & 50                               & 200                               \\
ara-yor            & 1397                               & 50                               & 100                               & fra-som                                    & 4231                               & 50                               & 200                               & swa-mlg                                    & 5000                               & 50                               & 200                               \\
bam-eng            & 4461                               & 50                               & 200                               & fra-swa                                    & 5000                               & 50                               & 200                               & swa-som                                    & 1390                               & 50                               & 173                               \\
bas-eng            & 5000                               & 50                               & 200                               & fra-swc                                    & 5000                               & 50                               & 200                               & swa-yor                                    & 24                                 & 3                                & 3                                 \\
bem-eng            & 5000                               & 50                               & 200                               & gez-amh                                    & 4619                               & 50                               & 200                               & swc-fra                                    & 5000                               & 50                               & 200                               \\
btg-fan            & 222                                & 28                               & 27                                & gez-eng                                    & 4724                               & 50                               & 200                               & teo-ach                                    & 5000                               & 50                               & 200                               \\
eng-aar            & 1146                               & 50                               & 143                               & hau-eng                                    & 5000                               & 50                               & 200                               & teo-eng                                    & 5000                               & 50                               & 200                               \\
eng-ach            & 5000                               & 50                               & 200                               & ibo-eng                                    & 5000                               & 50                               & 200                               & teo-lgg                                    & 5000                               & 50                               & 200                               \\
eng-afr            & 5000                               & 50                               & 200                               & kau-eng                                    & 4257                               & 50                               & 200                               & teo-lug                                    & 5000                               & 50                               & 200                               \\
eng-aka            & 3145                               & 50                               & 197                               & kbp-eng                                    & 5000                               & 50                               & 200                               & teo-nyn                                    & 5000                               & 50                               & 200                               \\
eng-amh            & 5000                               & 50                               & 200                               & kin-eng                                    & 5000                               & 50                               & 200                               & tir-aar                                    & 1165                               & 50                               & 145                               \\
eng-ara            & 5000                               & 50                               & 200                               & kon-eng                                    & 4357                               & 50                               & 200                               & tir-amh                                    & 5000                               & 50                               & 200                               \\
eng-bam            & 4462                               & 50                               & 200                               & lgg-ach                                    & 5000                               & 50                               & 200                               & tir-eng                                    & 5000                               & 50                               & 200                               \\
eng-bas            & 5000                               & 50                               & 200                               & lgg-lug                                    & 5000                               & 50                               & 200                               & tir-orm                                    & 4981                               & 50                               & 200                               \\
eng-bem            & 5000                               & 50                               & 200                               & lgg-nyn                                    & 5000                               & 50                               & 200                               & tir-som                                    & 1153                               & 50                               & 144                               \\
eng-ewe            & 128                                & 16                               & 16                                & lgg-teo                                    & 5000                               & 50                               & 200                               & tir-wal                                    & 2024                               & 50                               & 126                               \\
eng-gez            & 4724                               & 50                               & 200                               & lin-eng                                    & 246                                & 31                               & 31                                & tsn-eng                                    & 5000                               & 50                               & 200                               \\
eng-hau            & 5000                               & 50                               & 200                               & lin-fra                                    & 4000                               & 50                               & 200                               & tso-eng                                    & 5000                               & 50                               & 200                               \\
eng-ibo            & 5000                               & 50                               & 200                               & lug-ach                                    & 5000                               & 50                               & 200                               & twi-eng                                    & 5000                               & 50                               & 200                               \\
eng-kau            & 4257                               & 50                               & 200                               & lug-eng                                    & 5000                               & 50                               & 200                               & wal-amh                                    & 3763                               & 50                               & 200                               \\
eng-kbp            & 5000                               & 50                               & 200                               & lug-lgg                                    & 5000                               & 50                               & 200                               & wal-eng                                    & 5000                               & 50                               & 200                               \\
eng-kin            & 5000                               & 50                               & 200                               & lug-nyn                                    & 5000                               & 50                               & 200                               & wal-orm                                    & 2338                               & 50                               & 147                               \\
eng-kon            & 4357                               & 50                               & 200                               & lug-teo                                    & 5000                               & 50                               & 200                               & wal-tir                                    & 2024                               & 50                               & 127                               \\
eng-lin            & 246                                & 31                               & 31                                & mlg-amh                                    & 693                                & 43                               & 87                                & wol-eng                                    & 5000                               & 50                               & 200                               \\
eng-lug            & 5000                               & 50                               & 200                               & mlg-eng                                    & 5000                               & 50                               & 200                               & xho-eng                                    & 5000                               & 50                               & 200                               \\
eng-mlg            & 5000                               & 50                               & 200                               & mlg-swa                                    & 5000                               & 50                               & 200                               & yor-ara                                    & 1397                               & 50                               & 100                               \\
eng-nya            & 1052                               & 50                               & 132                               & mlg-yor                                    & 10                                 & 1                                & 1                                 & yor-eng                                    & 5000                               & 50                               & 200                               \\
eng-nyn            & 5000                               & 50                               & 200                               & nnb-fra                                    & 5000                               & 50                               & 200                               & yor-mlg                                    & 10                                 & 1                                & 1                                 \\
eng-orm            & 5000                               & 50                               & 200                               & nya-eng                                    & 1052                               & 50                               & 132                               & yor-swa                                    & 24                                 & 3                                & 3                                 \\
eng-pcm            & 1681                               & 50                               & 105                               & nyn-ach                                    & 5000                               & 50                               & 200                               & zul-eng                                    & 5000                               & 50                               & 200                  \\
\bottomrule
\end{tabular}%
}
\caption{Statistics of each language pair in our dataset.}\label{tab:data_statistics}
\end{table*}

\end{document}